\documentclass{article}

\usepackage{microtype}
\usepackage{graphicx}
\usepackage{subfigure}
\usepackage{booktabs} 

\usepackage{url}
\usepackage{amsmath,amsthm,amssymb,float}

\usepackage{hyperref}

\usepackage[accepted]{icml2019}

\icmltitlerunning{Efficient Exploration through Intrinsic Motivation Learning for Unsupervised Subgoal Discovery in Model-Free HRL}

\begin{document}

\twocolumn[
\icmltitle{Efficient Exploration through Intrinsic Motivation Learning for Unsupervised Subgoal Discovery in Model-Free Hierarchical Reinforcement Learning}



\icmlsetsymbol{equal}{*}

\begin{icmlauthorlist}
\icmlauthor{Jacob Rafati}{ucm}
\icmlauthor{David C.~Noelle}{ucm}
\end{icmlauthorlist}

\icmlaffiliation{ucm}{Electrical Engineering and Computer Science, University of California, Merced, CA, USA}

\icmlcorrespondingauthor{Jacob Rafati}{jrafatiheravi@ucmerced.edu}

\icmlkeywords{Reinforcement Learning, Exploration, Subgoal Discovery, Intrinsic Motivation}

\vskip 0.3in
]



\printAffiliationsAndNotice{}  

\begin{abstract}
Efficient exploration for automatic subgoal discovery is a challenging problem in Hierarchical Reinforcement Learning (HRL). In this paper, we show that intrinsic motivation learning increases the efficiency of exploration, leading to successful subgoal discovery. We introduce a model-free subgoal discovery method based on unsupervised learning over a limited memory of agent's experiences during intrinsic motivation. Additionally, we offer a unified approach to learning representations in model-free HRL.
\end{abstract}

\section{Introduction}
Model-free Reinforcement Learning (RL) algorithms, attempts to find an optimal policy through learning the values of agent's actions at any state by computing the expected future rewards without having access to a model of the environment \citep{SuttonRS:1988:TD}. To learn an efficient policy, the agent should balance its \emph{exploration}, i.e. visiting already observed rewarding states, with its \emph{exploration}, i.e. searching for better rewarding states. Exploitation is an strategy to maximize the expected reward on the one step, but exploration may lead to a greater \emph{return}, i.e. total rewards, in the long run \citep{RL-Book:Sutton:Barto:1998}.

One of the challenges that arise in RL in real-world problems is that the state space can be very large. This has classically been called the \emph{curse of dimensionality}. Non-linear function approximators coupled with reinforcement learning have made it possible to learn abstractions over high dimensional state spaces \citep{Rafati-Noelle:2015:CogSci,Rafati-Noelle:2017:CCN,Rafati-Noelle:2019:sparse-arXiv,DeepMind:Nature:2015}. 

Common approaches to exploration, such as the $\epsilon$-greedy method \citep{RL-Book:Sutton:Barto:1998}, are not sufficiently efficient in exploring the state space to succeed on large-scale complex problems with sparse delayed rewards feedback \citep{Bellemare2016UCB}. Exploration methods based on novelty detection \citep{Meyer1991,Achiam2017} and curiosity-driven learning \citep{pathak18largescale} have been particularly successful in sparse tasks but these methods typically require a generative or predictive model of the state transition probabilities , which can be difficult to train when the states space are very high-dimensional \citep{EX2,Bellemare2016UCB}. Learning representations of the value function is challenging for these tasks, since the agent receives an undiagnostic constant value, such as $r=0$ for most experiences. 

Hierarchical Reinforcement Learning (HRL) methods attempt to address the  issues of RL in sparse tasks \citep{Barto:2003:HRL,ENC-HRL:Hengst2010,dayan1992feudal,dietterich2000hierarchical} by learning to operate over different levels of \emph{temporal abstraction} \citep{Sutton:1999:Option,parr1997reinforcement,Krishnamurthy2016HierarchicalRL}. One approach to temporal abstraction involves identifying a set
of states that make for useful
\emph{subgoals}. This introduces a major open problem in HRL: that of
\emph{subgoal discovery}. The efficient exploration of the environment states has a direct effect on successful subgoal discovery. Some approaches to subgoal discovery maintain the value function in a large look-up table~\citep{Sutton:1999:Option,Goel:2003:Subgoal,Simsek:2005:subgoal,McGovern2001Subgoal},
and most of these methods require building the state transition graph,
providing a model of the environment and the agent's possible
interactions with it~\citep{Machado:2017:Laplacian,Simsek:2005:subgoal,Goel:2003:Subgoal}. Subgoal discovery problem is specifically challenging for the model-free HRL framework, since the agent does not have access to a model of the environment.

Once useful subgoals are discovered, an HRL agent should be able to learn the skills to attain those subgoals through the use of \emph{intrinsic motivation} --- artificially rewarding the agent for attaining selected subgoals \citep{Singh:2010:intrinsic-motivation,Vigorito:2010:intrinsic-motivation}. In such systems, knowledge of the current subgoal is needed to estimate future intrinsic reward, resulting in value functions that consider subgoals along with states \citep{vezhnevets2017feudal}. Such a parameterized universal value function, $q(s,g,a;w)$, integrates the value functions for multiple skills into a single function, taking the current subgoal, $g$, as an argument. Intrinsic motivation intend to provide a way for exploration while learning useful skills \citep{Barto:2003:HRL}. \cite{Bellemare2016UCB} has shown a connection between theoretical foundations of intrinsic motivation and the count-based exploration methods. However, the role of intrinsic motivation learning in efficient exploration of sparse tasks, its effect on learning representations in model-free HRL and its connection to the subgoal discovery problem are open research problems.     

It is important to note that \emph{model-free} HRL, which does not
require a model of the environment, still often requires the learning
of useful internal representations of states. Recently, \cite{Kulkarni:2016:Meta-Controller} has offered a model-free HRL approach, called Meta-controller and Controller framework, in order to integrate temporal abstraction and intrinsic motivation learning and successfully solved the first room of the Montezuma's Revenge. But, their method relied on a prior knowledge of the environment, including manual selection of \emph{interesting objects} as subgoals for intrinsic motivation learning. 

In our previous work, \cite{Rafati2019phd,Rafati-Noelle:2019:HRL-arXiv,Rafati-Noelle:2019:AAAI}, we have addressed major open problems in the integration of internal representation learning, temporal abstraction,
automatic subgoal discovery, and intrinsic motivation learning, all
within the model-free HRL framework We propose and implement efficient and general methods for subgoal discovery using unsupervised learning and anomaly (outlier) detection \citep{Rafati-Noelle:2019:SPiRL,Rafati-Noelle:2019:AAAI-KEG}. The proposed method do not require information beyond that which is typically collected by the agent during model-free reinforcement learning, such as a small memory of recent experiences {(agent trajectories)}. In our proposed approach for learning representations in model-free HRL, we were fundamentally constrained in three ways, by design. First, we remained faithful to a model-free reinforcement learning framework, eschewing any approach that requires the learning or use of an environment model. Second, we were devoted to integrating subgoal discovery with intrinsic motivation learning, and temporal abstraction. Lastly, we focused on subgoal discovery algorithms that are likely to scale to large reinforcement learning tasks. The result was a unified model-free HRL algorithm that incorporates the learning of useful internal representations of states, automatic subgoal
discovery, intrinsic motivation learning of skills, and the learning
of subgoal selection by a ``meta-controller''. We
demonstrated the effectiveness of this algorithm by applying it to a
variant of the rooms task, as well as the initial screen of the ATARI 2600 game called
\emph{Montezuma's Revenge}.

In this paper, we investigate the role of intrinsic motivation learning on efficient exploration of environments with sparse delayed rewards feedback, and its connection to the subgoal discovery problem in our model-free HRL framework. We introduce an efficient and general method for subgoal discovery using unsupervised learning methods, such as K-means clustering and anomaly detection, over a small memory of agent's experiences (trajectories) during intrinsic motivation learning. Finally, we conjecture that intrinsic motivation learning can increase appropriate state space coverage, and it produces a policy for efficient exploration that leads to a successful subgoal discovery. We demonstrate the effectiveness of our method on the rooms task (Figure \ref{fig:rooms}(a)), as well as the initial screen of the \emph{Montezuma's Revenge} (Figure \ref{fig:montezuma}(a)). 

\section{A Model-Free HRL Framework}
\subsection{Meta-controller/Controller Framework}
Inspired by Kulkarni et al.~(\citeyear{Kulkarni:2016:Meta-Controller}) we start by using two levels of hierarchy for temporal abstraction learning (Figure \ref{fig:unified-hrl-framework}). The more abstract level of this hierarchy is managed by a \emph{meta-controller} which guides the action selection processes of the lower level \emph{controller}. This approach leads to integration of temporal abstraction and intrinsic motivation learning in deep model-free HRL framework. Separate value functions are learned for the meta-controller and the controller. 

At time step $t$, the meta-controller receives a state observation,
$s=s_t$, from the environment. It then selects
a \emph{subgoal}, $g=g_t$, from a set of subgoals, $\mathcal{G}$ from an $\epsilon$-greedy policy derived from the meta-controller's value function, $Q(s,g;\mathcal{W})$. With the current subgoal selected, the controller uses its policy to select an action, $a \in \mathcal{A}$, based on the current state, $s$, and the current subgoal, $g$. We used $\epsilon$-greedy policy derived from the controller's value function, $q(s,g,a;w)$, to choose an action $a$. In next time step, agent recieves a reward $r_{t+1}=r$, and the next state, $s_{t+1}=s'$, and stores its direct experiences with the environment into an experience memory, $\mathcal{D}$, Actions continue to be selected by the controller while an internal critic monitors the current state, comparing it to the current subgoal, and delivering an  appropriate \emph{intrinsic reward}, $\tilde{r}$, to the controller on each time step. Each transition experience, $(s,g,a,\tilde{r},s')$, is recorded in the controller's experience memory set, $\mathcal{D}_{1}$, to support learning. When the subgoal is attained, or a maximum amount of time has passed, the meta-controller observes the resulting state, $s_{t'}=s_{t+T+1}$, and selects another subgoal, $g'$. The transition experience for the meta-controller, $(s,g,G,s_{t'})$ is recorded in the meta-controller's experience memory set, $\mathcal{D}_{2}$, where $G=\sum_{t'=t}^{t+T} \gamma^{t'-t} r_{t'}$ is the return between the selection of consecutive
subgoals. For training the
meta-controller value function, $Q(s,g;\mathcal{W})$, we minimize its loss function based on
the experience received from the environment, $\mathcal{D}_2$. The controller improves its subpolicy, $\pi(a|s,g)$, by learning its value function, $q(s,g,a;w)$,
through minimization of its loss function over intrinsic experiences, $\mathcal{D}_{1}$. 
 
\subsection{Unsupervised Subgoal Discovery}
\label{sec:USD}
The performance of the meta-controller/controller framework depends
critically on selecting good candidate subgoals for the
meta-controller to consider. In \cite{Kulkarni:2016:Meta-Controller}'s approach to model-free HRL, the subgoals are manually specified for each task, and hence, the subgoal discovery in a model-free HRL framework and its contribution to learning representations of the meta-controller and controller value functions have not been addressed. In order to integrate the automatic subgoal discovery into the meta-controller/controller framework in model-free HRL, we should only use the information available from the intrinsic motivation learning.  Our strategy for subgoal discovery involves applying unsupervised learning methods to a recent experience memory, $\mathcal{D}$, to identify sets of states that may be good subgoal candidates. We focus specifically on two kinds of analysis that can be performed on the set of transition experiences. We hypothesize that good subgoals might be found by (1) attending to the states associated with \emph{anomalous} transition experiences and (2) clustering experiences based on a similarity measure and collecting the set of associated states into a potential subgoal. Thus, our proposed method merges \emph{anomaly (outlier) detection} with the $K$-means clustering of experiences.

The anomaly (outlier) detection process identifies states associated
with experiences that differ significantly from the others. In the
context of subgoal discovery, a relevant anomalous experience would be
one that includes a substantial positive reward in an environment in
which reward is sparse. For example, in the rooms task, transitions that arrive at the key or the lock are quite dissimilar to most transitions, due to the large positive reward
that is received at that point (see Figure \ref{fig:rooms} (b-d)). Similarity in the Montezuma's Revenge game, action that lead to or the \textit{doors} lead to rewarding experiences. Additionally, when agent enters a new room, the state becomes very different than the states from the previous room, allowing an anomaly detection method to identify the door that leads to a new room as a potentially useful subgoal.    

The idea behind using the clustering of experiences involves both ``spatial'' state space abstraction and dimensionality reduction with regard to the internal representations of states. The learning process
might be made faster by considering representative states, such as cluster centroids as candidate subgoals, rather than considering all the states. For example, in the
rooms task, the centroids of $K$-means clusters, with $K=4$ (Figure \ref{fig:rooms} (b)), lie close to the geometric center of each room, with the states within each room coming to belong to the corresponding subgoal's cluster. In this way, the clustering of transition experiences can approximately produce a coarser representation of state space, in this case replacing the fine grained ``grid square location'' with the coarser
``room location''. Also, $K$-means clustering found useful subgoal regions, such as ladders, stages, and the rope in Montezuma's Revenge game (Figure \ref{fig:montezuma}(c)).

\subsection{A Unified Approach to Model-Free HRL}
\begin{figure}[htb!]
	\centering
	\includegraphics[width=0.22\textwidth]{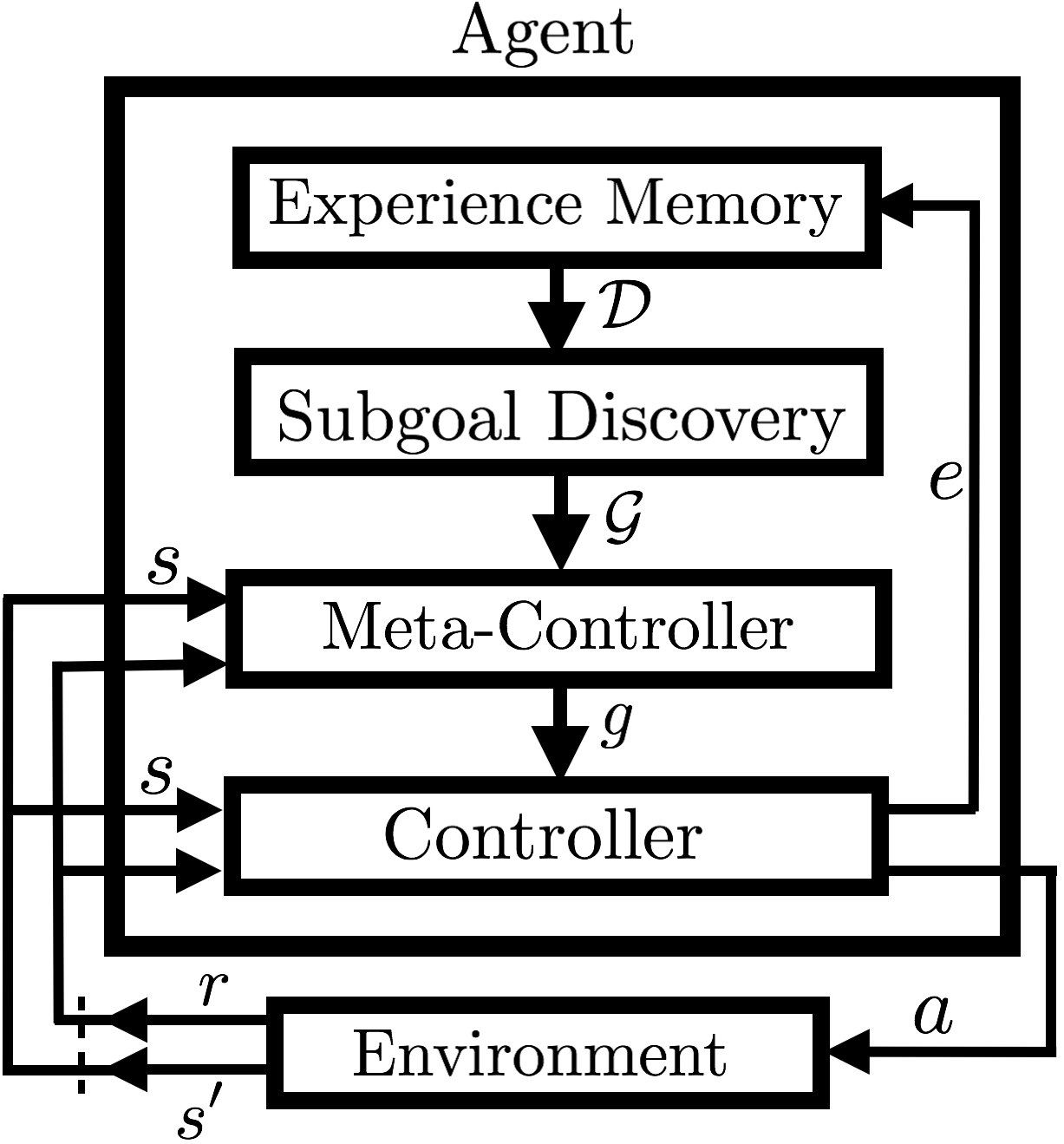}
	\caption{(a) The information flow in our unified model-free hierarchical reinforcement learning framework.}
	\label{fig:unified-hrl-framework}
\end{figure}

Here, we offer a framework to integrate learning representations of the meta-controller and controller value functions, and also unsupervised subgoal discovery for model-free HRL approach. The agent's experience memory, $\mathcal{D}$, which has been called \textit{experience replay memory} in deep Q-learning, is the necessary element for integrating the HRL components into a unified framework. The information flow between the components of our unified model-free HRL is depicted in Figure \ref{fig:unified-hrl-framework}. We applied this method to a variant of rooms tasks with a key and a box (Figure \ref{fig:rooms} (a)), and also a first screen of the Montezuma's Revenge ATARI game (Figure \ref{fig:montezuma}(a)). In these simulations, learning occurred in one unified phase. The meta-controller and the controller, and unsupervised subgoal discovery, were trained all together. The average return for the unified HRL method, and regular RL is shown in Figure \ref{fig:rooms}(f). In Montezua's Revenge, the controller was initially trained to navigate the man in red to random subgoals on the screen, derived from the Canny edge detection algorithm (see Figure \ref{fig:montezuma}(b)). Using this strategy, the agent learned navigation skills, detected the rewarding states: \emph{key} and \emph{doors}, and other interesting regions. Our model-free HRL could solved this room, while deep Q-learning networks \citep{DeepMind:Nature:2015} could not.

\subsection{Intrinsic Motivation for Efficient Exploration}
\begin{figure*}[htb!]
	\centering
	\begin{tabular}{cccc}
		\includegraphics[width=0.22\textwidth]{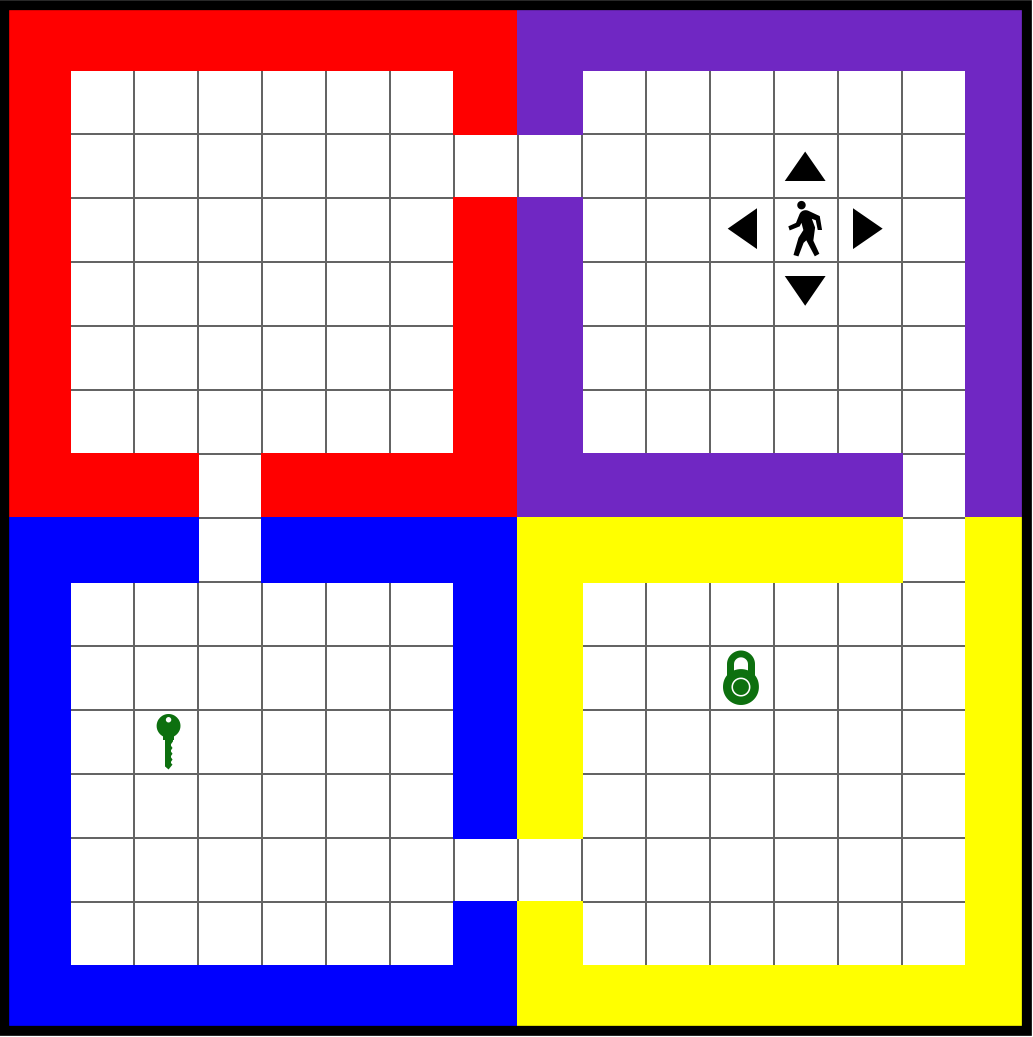} &
		\includegraphics[width=0.22\textwidth]{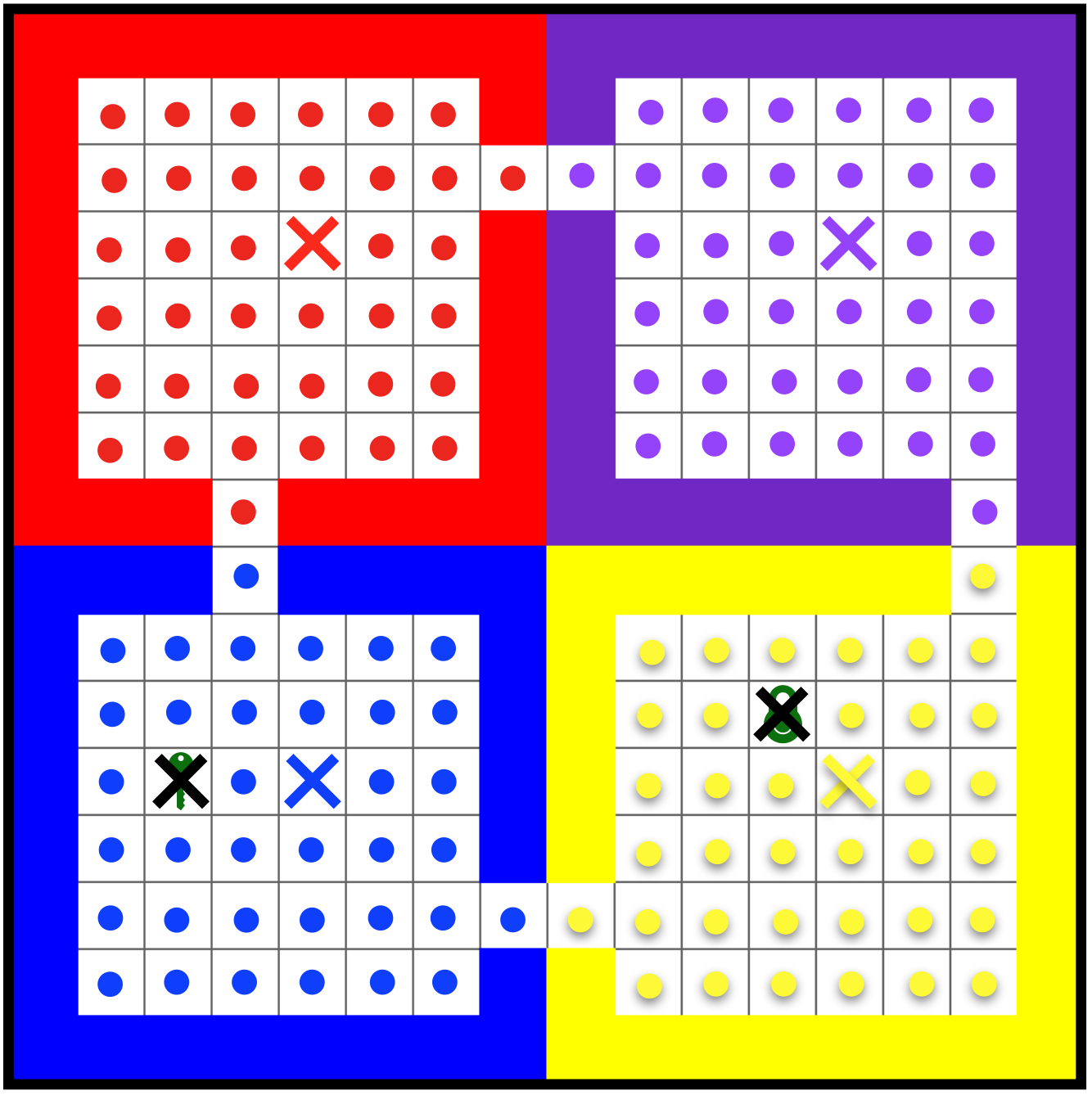} &
		\includegraphics[width=0.22\textwidth]{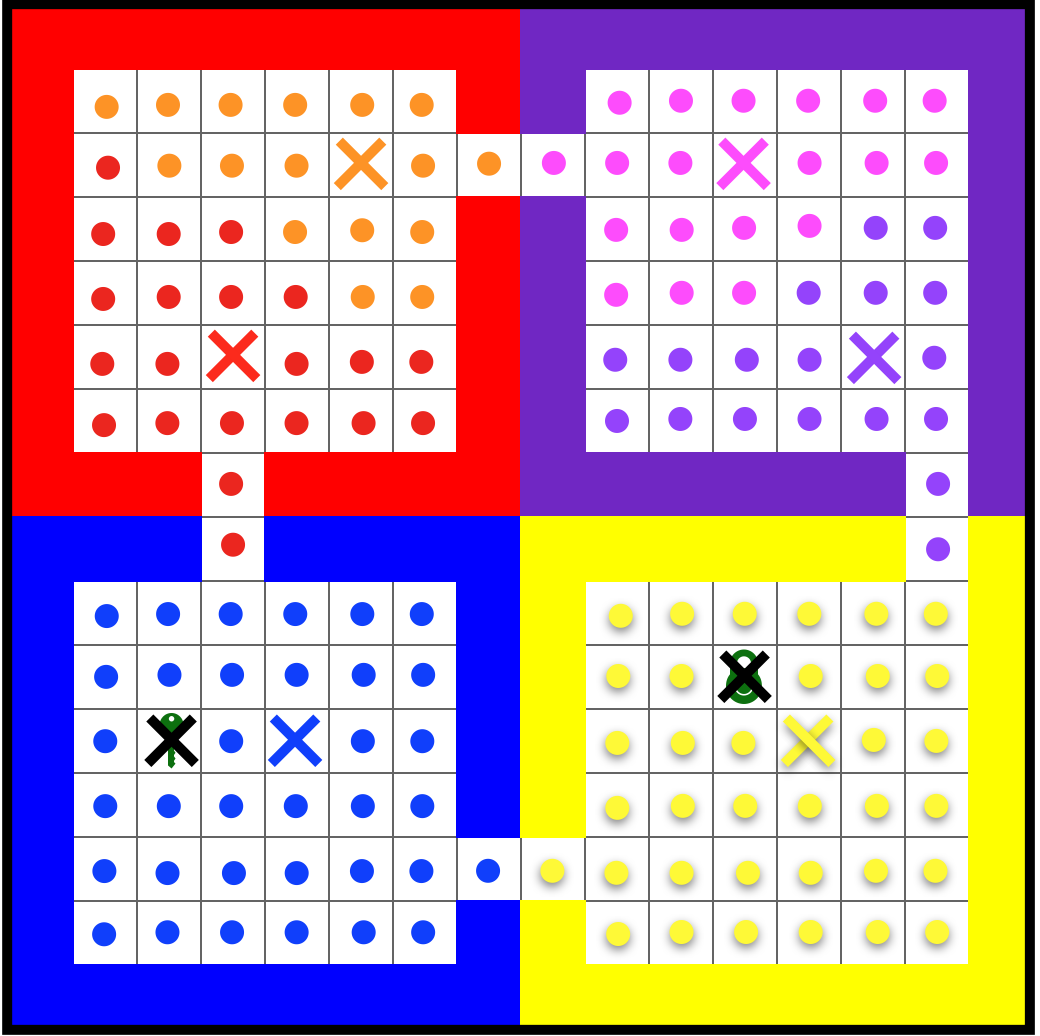} &
		\includegraphics[width=0.22\textwidth]{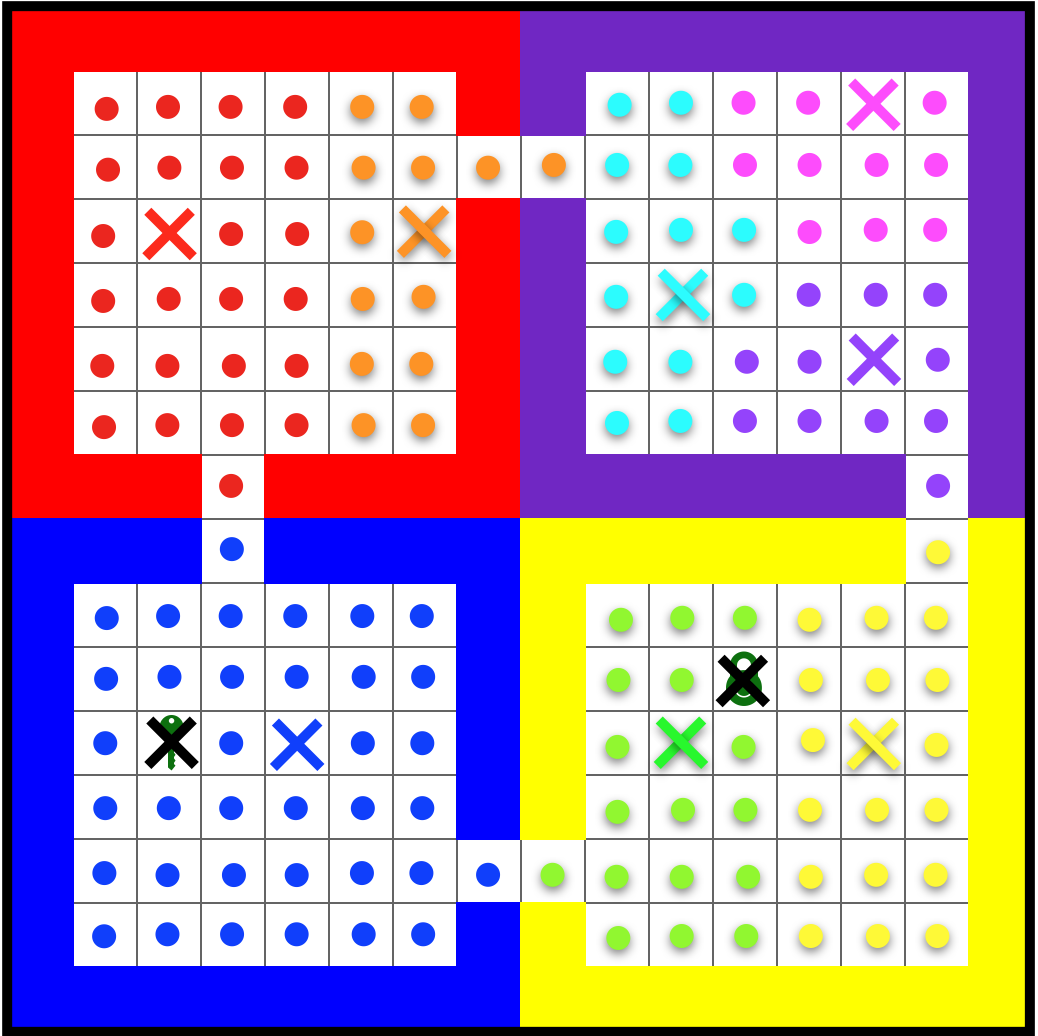}\\
		(a) & (b) & (c) & (d)\\
		\multicolumn{2}{c}{\includegraphics[width=0.35\textwidth]{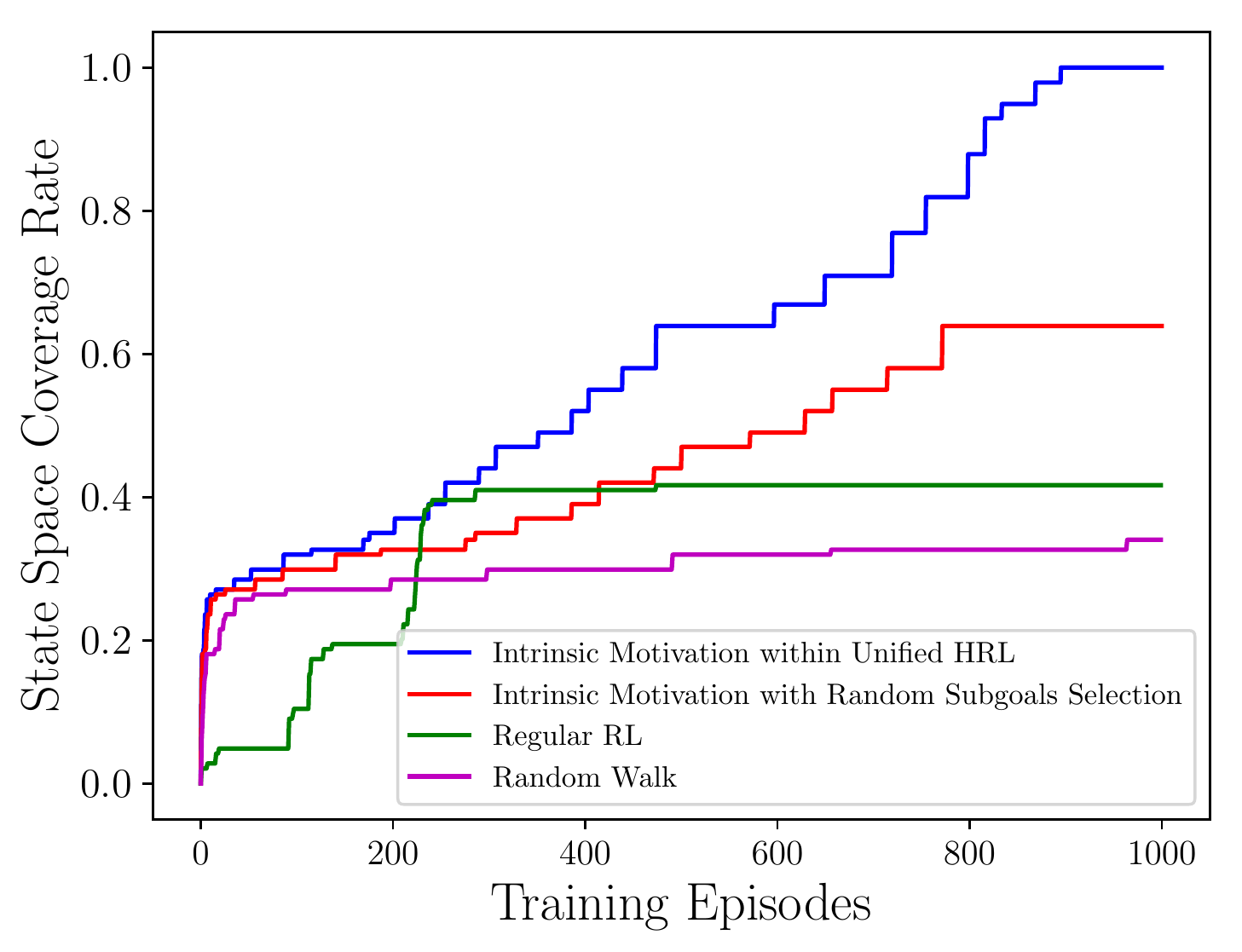}} & \multicolumn{2}{c}{\includegraphics[width=0.35\textwidth]{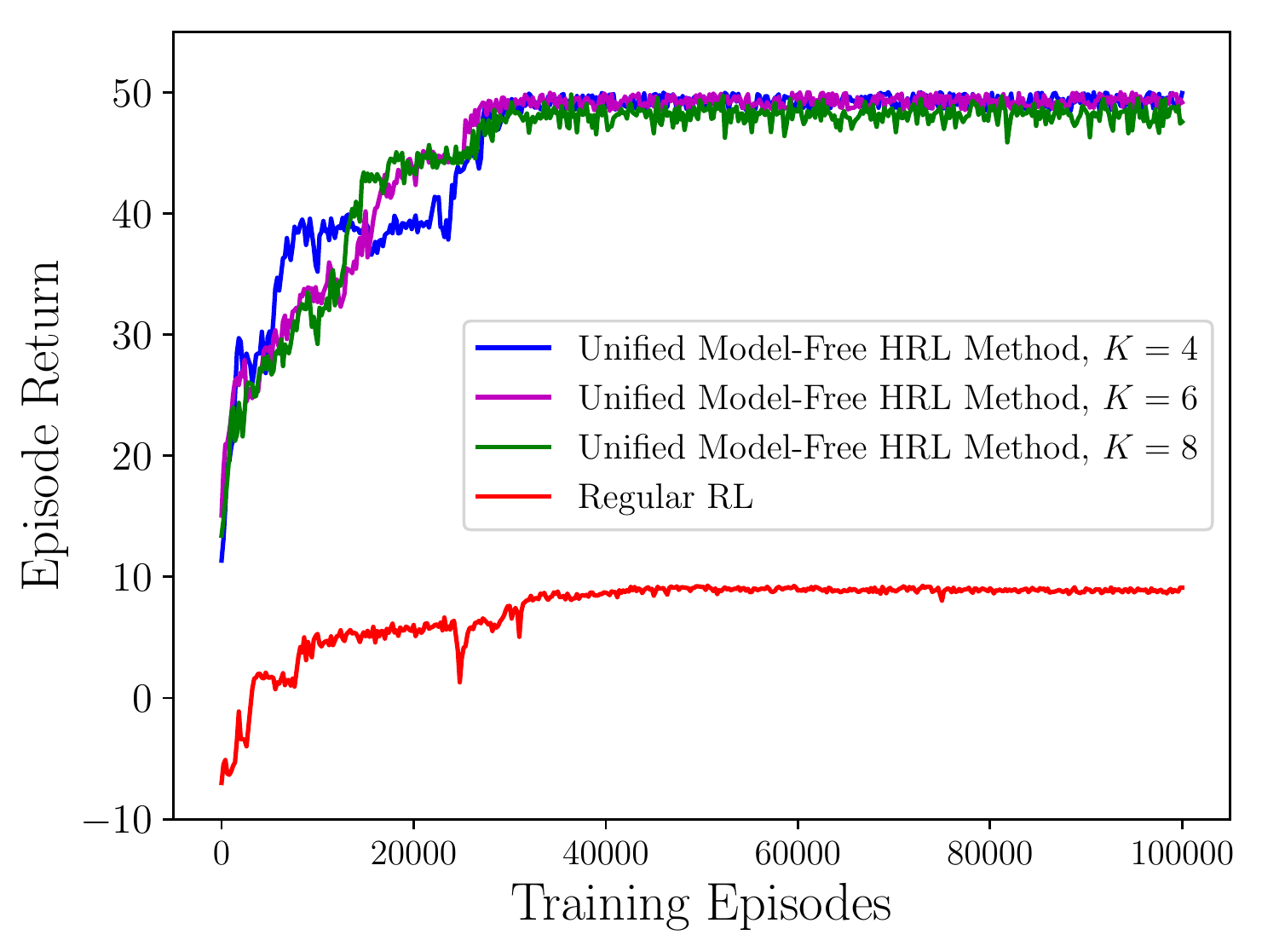}}\\
		\multicolumn{2}{c}{(e)} & \multicolumn{2}{c}{(f)}
	\end{tabular}
	\caption{(a) The \emph{4-room} task with a {key} ($r=+10$) and a {lock} ($r=+40$). (b-d) The results of the unsupervised subgoal discovery algorithm with \emph{anomalies} marked with black Xs and \emph{centroids} with colored ones. The number of $K$-means clusters was set to (b) $K=4$, (c) $K=6$,  (d) $K=8$. (e) The rate of visited number of states to the total states. (f) The average episode return.}
	\label{fig:rooms}
\end{figure*}

\begin{figure*}[htb!]
	\centering
	\begin{tabular}{cccc}
		\includegraphics[width=0.16\textwidth]{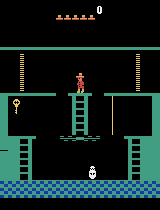} &
		 \includegraphics[width=0.16\textwidth]{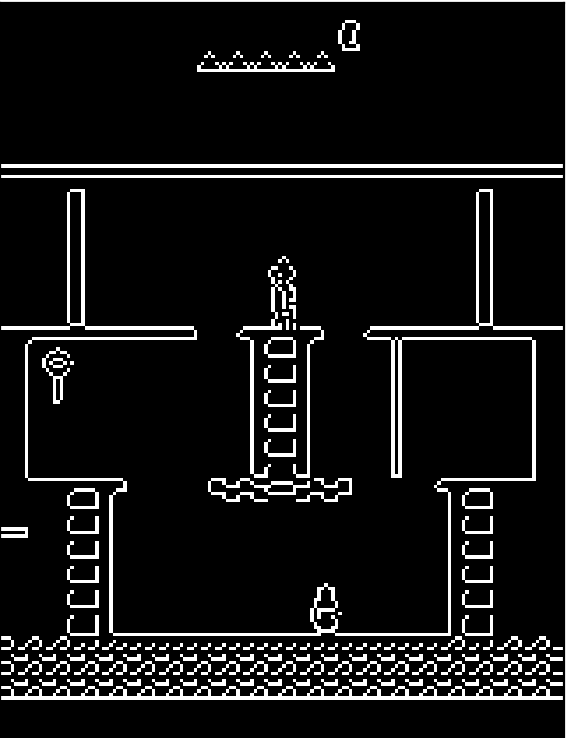} & \includegraphics[width=0.16\textwidth]{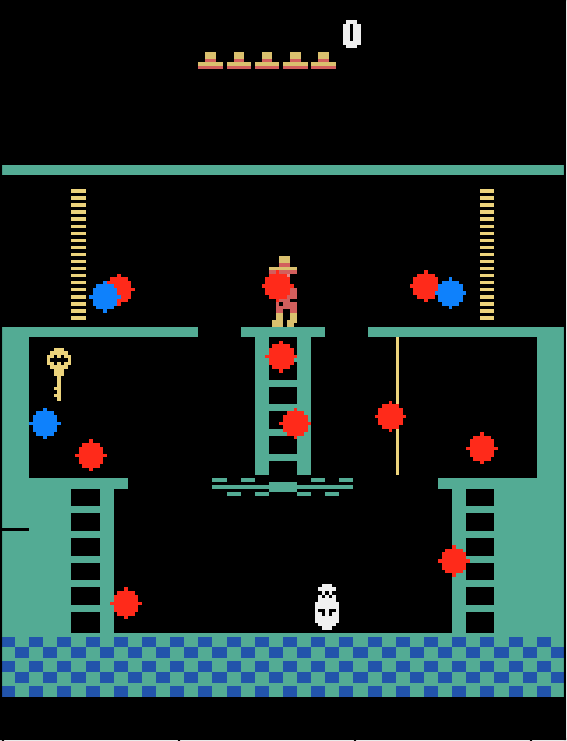} & 
		 \includegraphics[width=0.16\textwidth]{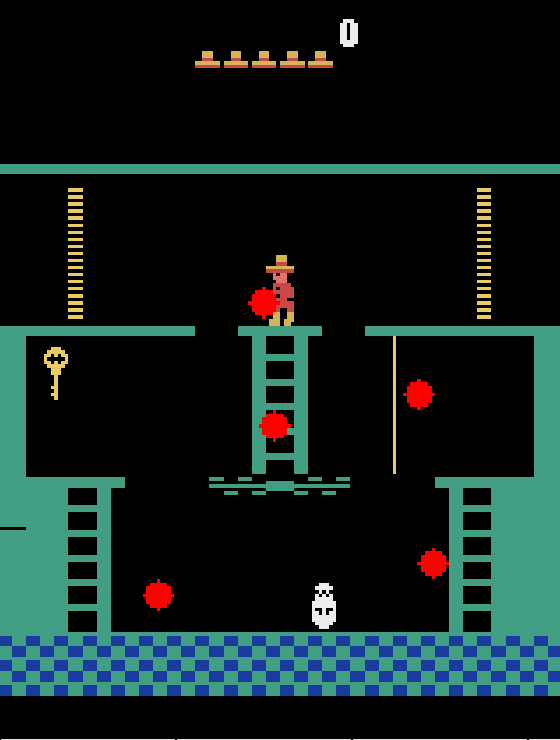} \\
		 (a) & (b) & (c) & (d) \\
		\multicolumn{2}{c}{\includegraphics[width=0.35\textwidth]{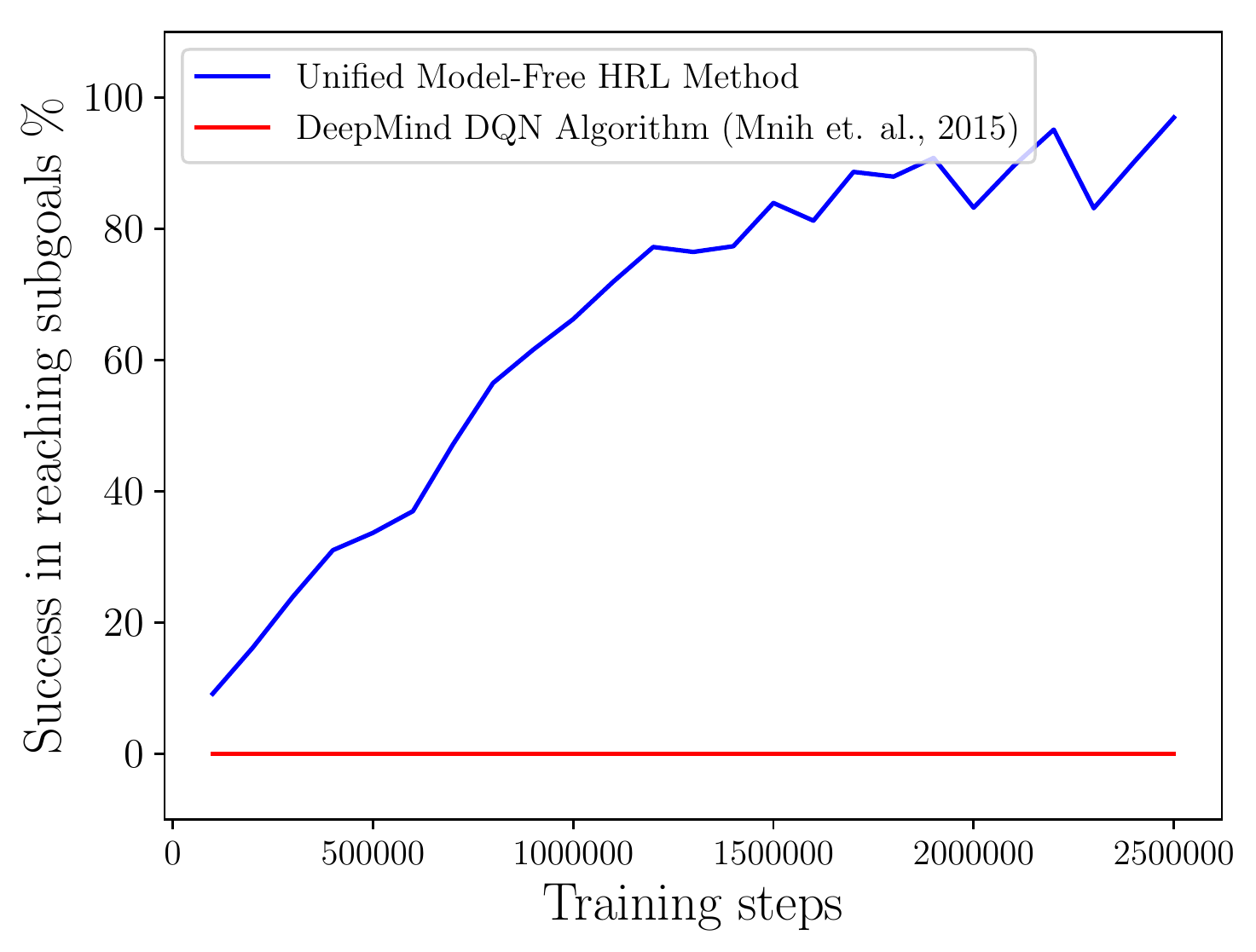}} & 
		\multicolumn{2}{c}{\includegraphics[width=0.35\textwidth]{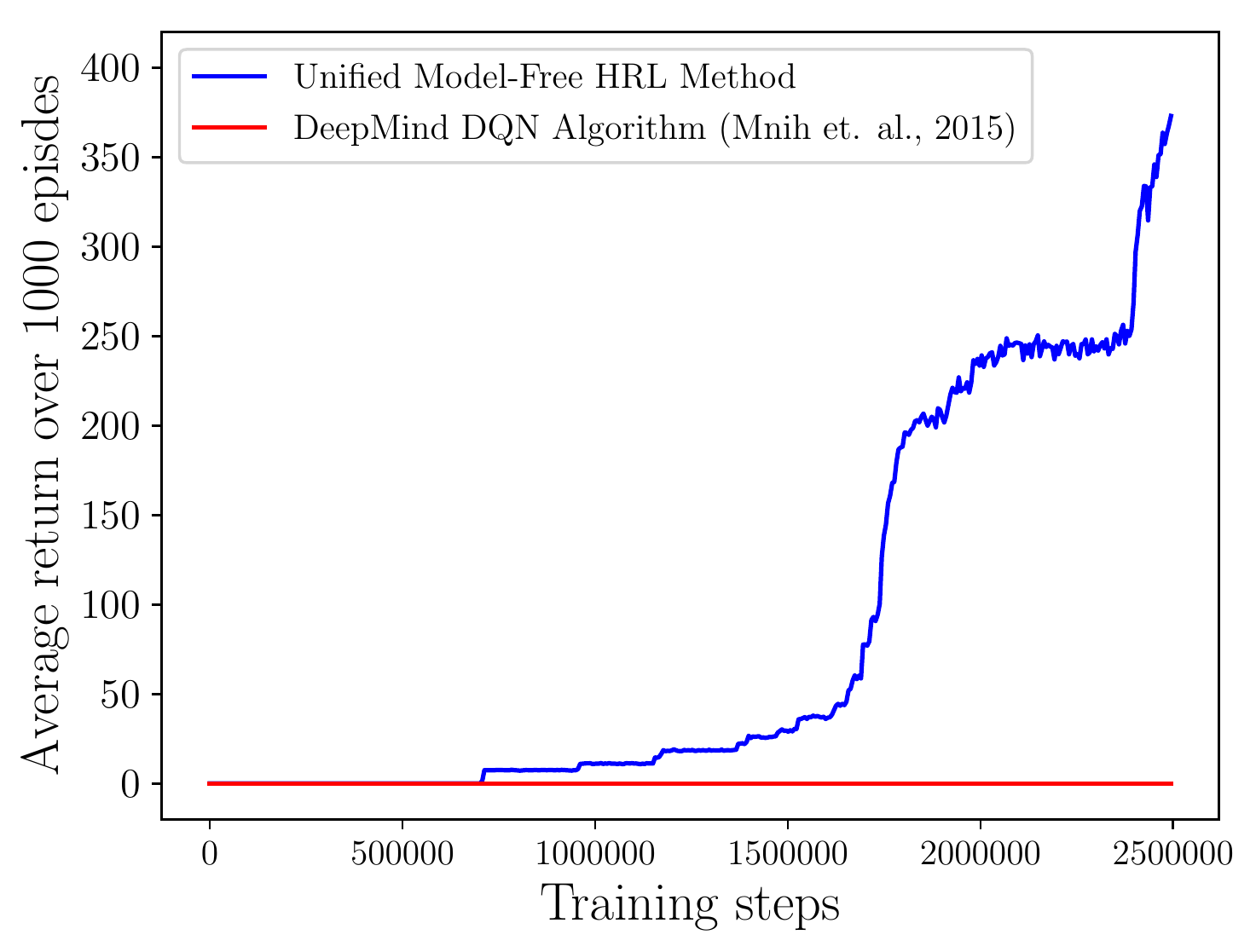}}\\
		\multicolumn{2}{c}{(e)} & \multicolumn{2}{c}{(f)}
	\end{tabular}
	\caption{(a) The first screen of the Montezuma's Revenge game. (b) The results of an off-the-shelf Canny edge detection on a single image of the game. (c) The results of the unsupervised subgoal discovery algorithm during intrinsic motivation learning in the first room of Montezuma's Revenge. Blue circles are the anomalous subgoals and the red ones are the centroids of clusters. (d) The results of the unsupervised subgoal discovery for a random walk. (e) The success of the controller in reaching subgoals. (f) The average game score. }
	\label{fig:montezuma}
\end{figure*}

Intrinsic motivation learning is the core idea behind the learning of
the value function for the controller. The intrinsic critic in this HRL framework can send much more regular feedback to the controller, since it is based on attaining subgoals, rather than ultimate goals. In our unified model-free HRL, the intrinsic motivation learning plays two major roles: (1) Learning skills to go from any observable states to other region of states through learning subpolices by the controller.
(2) Providing efficient exploration to collect experiences that can be used for unsupervised subgoal discovery. The state space coverage rate, i.e. the number of visited states divided by the size of states space during training rooms tasks is shown in Figure \ref{fig:rooms}(e). Intrinsic motivation learning coupled with unsupervised subgoal discovery and a random meta-controller can visit 67\% of the states. A regular Q-learning method converges to a solution that can finds and picks the \textit{key}, but it doesn't have motivation to explore other regions to find more rewarding state, i.e. the \textit{box}. When intrinsic motivation learning of the controller is integrated with unsupervised subgoal discovery and a meta-controller, the unified method can successfully cover 100\% of the states. A random walk in Montezuma's Revenge can only visit two ladders and the rope (see Figure \ref{fig:montezuma} (d)). But intrinsic motivation learning with unsupervised subgoal discovery can lead to discovery of all meaningful regions of the screen including rewarding ones (Figure \ref{fig:montezuma}(c)). 

\section{Conclusions}
We introduced a novel method for subgoal discovery, using unsupervised learning over a small memory of the most recent experiences of the agent. Intrinsic motivation learning provides a policy for efficient exploration of sparse tasks that leads to successful unsupervised subgoal discovery. We investigated the role of the intrinsic motivation learning on  efficient exploration of the observable state space for discovering useful subgoals. 

\bibliography{refs}
\bibliographystyle{icml2019}

\end{document}